\documentclass[preprint,authoryear,12pt]{elsarticle}

\usepackage{graphicx}% Include figure files

\begin{document}
\begin{frontmatter}

\title{Measures of lexical distance between languages}
\author{Filippo Petroni}
\address{DIMADEFA, Facolt\`a di Economia, 
Universit\`a di Roma "La Sapienza", 
I-00161 Roma, Italy\\email: fpetroni@gmail.com}
\author{Maurizio Serva}
\address{Dipartimento di Matematica,
Universit\`a dell'Aquila,
I-67010 L'Aquila, Italy\\email: serva@univaq.it}

\begin{abstract}
The idea of measuring distance between languages seems to have its 
roots in the work of the French explorer Dumont D'Urville \cite{Urv}. 
He collected comparative words lists of various languages during his 
voyages aboard the Astrolabe from 1826 to 1829 and, in his work about 
the geographical division of the Pacific, he proposed a method to measure 
the degree of relation among languages. The method used by modern 
glottochronology, developed by Morris Swadesh in the 1950s,
measures distances from the percentage of shared cognates, which are words 
with a common historical origin.  Recently, we proposed a new automated method
which uses normalized Levenshtein distance among words with 
the same meaning and averages on the words contained in a list.
Recently another group of scholars \cite{Bak, Hol}
proposed a refined of our definition
including a second normalization.
In this paper we compare the information content of our definition
with the refined version in order to decide which of the two
can be applied with greater success to
resolve relationships among languages.
\end{abstract}

%\pacs{87.23.Ge; 87.23.Kg; 89.75.Hc}

\end{frontmatter}
 
\section{Introduction}
Glottochronology tries to estimate the time at which languages diverged 
with the implicit assumption that vocabularies change at a constant average
rate. The idea is to consider the percentage of shared cognates in order 
to compute the distance between pairs of languages \cite{Sw}.  
These  lexical  distances are assumed to be, on average, 
logarithmically proportional to divergence times.  
In fact, changes in vocabulary accumulate year after year and two 
languages initially similar become more and more different. 
A recent example of the use of Swadesh lists and cognates to construct 
language trees are the studies of Gray and Atkinson \cite{GA}
and Gray and Jordan \cite{GJ}.

We recently proposed an automated method which uses Levenshtein distance 
among words in a list \cite{SP, SP2}. 
To be precise, we defined the lexical distance of two languages  
by considering a normalized Levenshtein distance among words with 
the same meaning and averaging on all the words contained in a Swadesh list. 
The normalization is extremely important
and no reasonable results can be found without.
Then, we transformed the lexical 
distances in separation times.
This goal was reached by a logarithmic rule which
is the analogous of the adjusted fundamental formula of 
glottochronology \cite{ST}.
Finally, the phylogenetic tree could be straightforwardly constructed.

In \cite{SP, SP2} we tested our method by
constructing the phylogenetic trees of the Indo-European  
and the Austronesian groups.

Almost at the same time, the above described
automated method was used and developed 
by another large group of scholars \cite{Bak, Hol}. 
They placed the method at the core of an ambitious
project, the ASJP (The Automated Similarity Judgment Program).
In their work they proposed a refined of our definition
including a second normalization in the definition
of lexical distance.

The goal of this paper is to compare 
the information content of the two definitions
in order to decide which of the two
can be applied with greater success to
resolve relationships among languages.

Before tackling this problem we sketch our definition
of lexical distance and the modification proposed
in \cite{Bak, Hol} which is a refinement including a second normalization.
Then we compare the information content of the two definitions and give our
conclusion.

\section{Lexical distance}
Our definition of lexical distance between two words is 
a variant of the Levenshtein distance which is simply the minimum 
number of  insertions, deletions, or substitutions of a 
single character needed to transform one word into the other.
Our definition is taken as the Levenshtein distance divided  by the 
number of characters of the longer of the two compared words.
More precisely, given two words $\alpha_i$ and $\beta_j$
their lexical distance $D(\alpha_i, \beta_j)$ is given by

\begin{equation}
D(\alpha_i, \beta_j)= 
\frac{D_l(\alpha_i, \beta_j)}{L(\alpha_i, \beta_j)}
\label{wd}
\end{equation}
where $D_l(\alpha_i, \beta_j)$ is the 
Levenshtein distance between the two words
and $L(\alpha_i, \beta_j)$ is the
number of characters of the longer of the two words
$\alpha_i$ and $\beta_j$.
Therefore, the distance can take any value between 0
and 1. Obviously  $D(\alpha_i, \alpha_i)=0$ .

The normalization is an important novelty and it
plays a crucial role;
no sensible results can been found without\cite{SP, SP2}.

We use distance between pairs of words, as defined above, 
to construct the lexical distances of languages. 
For any pair of languages, the first step is to compute
the distance between  words corresponding to the same meaning 
in the Swadesh list. 
Then, the lexical distance between each
languages pair is  defined as the average of the 
distance between all words\cite{SP, SP2}.  
As a result we have a number between 0 and 1 which 
we claim to be the lexical distance between two languages. 
 
Assume that the number of languages is $N$
and the list of words for any language contains
$M$ items. 
Any language in the group is labeled a Greek letter
(say $\alpha$) and any word of that language by 
$\alpha_i$ with $1 \leq i \leq M$. 
Then, two words $\alpha_i$ and $\beta_j$  
in the languages $\alpha$ and $\beta$ 
have the same meaning if $i=j$.

Then the distance between two languages is

\begin{equation}
D(\alpha, \beta)=  \frac{1}{M} \sum_i
D(\alpha_i, \beta_i)
\label{ld}
\end{equation}
where the sum goes from 1 to $M$.
Notice that only pairs of words with same meaning are 
used in this definition. This number
is in the interval [0,1], obviously $D(\alpha, \alpha)=0$.

The results of the analysis is a   
$N \times N $ upper triangular matrix
whose entries are the $N(N-1)$ non trivial lexical 
distances $D(\alpha, \beta)$ between all pairs in a group.
Indeed, our method for computing distances is a very 
simple operation, 
that does not need any specific linguistic knowledge 
and requires a minimum of computing time.

A phylogenetic tree could be constructed from 
the matrix of lexical distances $D(\alpha, \beta)$,
but this would only give the topology of the tree
whereas the absolute time scale would be missing.
Therefore, we perform \cite{SP, SP2} a logarithmic transformation 
of lexical distances which is the analogous of
the adjusted fundamental formula of glottochronology\cite{ST}.
In this way we obtain a new
$N \times  N$ upper triangular matrix whose entries
are the divergence times
between all pairs of languages.  
This matrix preserves the topology of the lexical 
distance matrix but it also contains the information 
concerning absolute time scales.
Then, the phylogenetic tree can be straightforwardly constructed.

In \cite{SP, SP2} we tested our method 
constructing the phylogenetic trees of the Indo-European group 
and of the Austronesian group.
In both cases we considered $N=50$ languages.
The database\cite{footnote}  that we used in
\cite{SP, SP2}  is 
composed by $M=200$ words for any language.
The main source for the database for the Indo-European group 
is the file prepared by Dyen et
al. in \cite{D}.  
For the Austronesian group we used as the main source the 
lists contained in the huge database in \cite{NZ}.

\section{A second normalization}
A further modification has been proposed by
\cite{Bak, Hol} in order
to avoid possible similarity which 
could arose from accidental relative orthographical
similarity of languages.

Let us first define the {\it global distance} 
between languages $\alpha$ and $\beta$  as

\begin{equation}
\Gamma(\alpha, \beta)=  \frac{1}{M(M-1)} \sum_{i \neq j}
D(\alpha_i, \beta_j)
\label{gd}
\end{equation}
where the sum goes on all $M(M-1)$ pairs of
words corresponding to different
meanings in the two lists
($M^2$ is the total number of pairs and
$M$ is the number of pairs with same meaning).

This quantity measures a distance of the vocabulary of the two languages, 
without comparing words with same meaning. In other words,
it only account for general similarities
in the frequency and ordering of characters.
The point is that, at this stage, we don't know if
$\Gamma(\alpha, \beta)$ carries informations
or only depends on accidental similarities.

Assuming the second point of view,
it is reasonable to define, according to \cite{Bak, Hol},
a bi-normalized lexical distance as follows:

\begin{equation}
D_s(\alpha, \beta)=  
\frac{D(\alpha, \beta)}{\Gamma(\alpha, \beta)}
\label{lds}
\end{equation}
This second normalization should cancel
the effects of accidental orthographical
similarities between the two languages.
Notice that while by definition $D(\alpha, \alpha)=
D_s(\alpha, \alpha)=0$,
in all real cases $\Gamma(\alpha, \alpha) \neq 0$.

We would like to stress that the idea of the proposed second 
normalization turns to be correct only if $\Gamma(\alpha, \beta)$
is uncorrelated with the lexical distance 
between languages $\alpha$ and $\beta$.
In this case, in fact, it has vanishing information concerning
their relationship.
On the contrary, if it is positively
correlated with the distance between the
two languages, one can conclude that 
it contains some information that can be usefully exploited.

\section{Comparison of different definitions}
In order to decide which definition is better to use, 
$D(\alpha,\beta)$ or $D_s(\alpha, \beta)$,
we have to see if $\Gamma(\alpha,\beta)$ is
positively correlated with these distances.
In case it is not, we will decide to
use $D_s(\alpha, \beta)$
since we eliminate errors due to
accidental similarities between vocabularies.
On the contrary, if it is positively correlated,
we would conclude that $\Gamma(\alpha,\beta)$ carries
some positive information about the degree 
of similarity of the two languages.
In this second case, two languages will be, in average,
closer for smaller $\Gamma(\alpha,\beta)$ and
we would decide to use
$D(\alpha, \beta)$ since it incorporates
the information contained in $\Gamma(\alpha,\beta)$.

In order to compute the correlation between distance 
and $\Gamma(\alpha,\beta)$ we proceed as follows:
first we define for a generic function 
$ f(\alpha, \beta)$ the average $<$$f$$>$ on all possible  
values of $\alpha$ and $\beta$ as follows

\begin{equation}
<\!\!f\!\!> = \frac{1}{N^2} \sum_{\alpha, \beta}
f(\alpha, \beta)
\label{mean}
\end{equation}
which is the average value
of the function $f(\alpha, \beta)$ in a 
linguistic group.
Then, we define the correlation between
$D(\alpha, \beta)$ and $\Gamma(\alpha, \beta)$
in a standard way as

\begin{equation}
C(\Gamma,D)= \frac{<\!\!(\Gamma -<\!\!\Gamma\!\!>) 
(D-<\!\!D\!\!>)\!\!>}
{(<\!\!(\Gamma- <\!\!\Gamma\!\!>)^2\!\!>
<\!\!(D- <\!\!D\!\!>)^2\!\!>)^{\frac{1}{2}}} 
\label{c}
\end{equation}

The result is that the correlation in the 
Indo-European group is   $0.59173$
while in the Austronesian group is   $0.46032$.
In both cases it is a quite high positive value
(correlation may take any value between -1 and 1)
and we conclude that eventual vocabulary 
similarities accounted by $\Gamma(\alpha, \beta)$ 
carry information and are not at all accidental. 
The week point is that we have checked correlation 
against $D(\alpha, \beta)$ which,
at least from the point of view of the proponents
of the second normalization, linearly incorporates
$\Gamma(\alpha, \beta)$ since
$D(\alpha, \beta) =\Gamma(\alpha, \beta) 
D_s(\alpha, \beta)$. 

From this point of view
our result is not so astonishing.   
Nevertheless, we can also compute the
correlation between the bi-normalized distance
$D_s(\alpha, \beta)$ and $\Gamma(\alpha, \beta)$.
The definition is the same as (\ref{c})
with $D_s$ substituting $D$.
We obtain that the correlation 
$C(\Gamma,D_s) $  in the 
Indo-European group is $0.54713$
while in the Austronesian group is $0.40169$.
These two data, although slightly smaller than the previous 
ones, are still quite high and confirm that
$\Gamma(\alpha, \beta)$ contains positive information.
In other words, closer languages, both
in the sense of a smaller $D(\alpha, \beta)$ and a 
smaller $D_s(\alpha, \beta)$, will have on average
smaller $\Gamma(\alpha, \beta)$. 

We remark that the same correlation coefficients,
both for $D$ and $D_s$, comes out,
if the average (\ref{mean}) is computed
negletting the pairs were the same greek index is repeated.

In order to complete our analysis 
we plot, only for the Austronesian languages group,
$\Gamma(\alpha, \beta)$ as a function
of $D(\alpha, \beta)$  (Fig. 1 left)
and as a function of $D_s(\alpha, \beta)$  (Fig. 1 right).
Any point in the figures represents a pair of 
languages.
In both cases we perceive the positive
correlation which is evidenced by the best 
linear fits. 

We remark that the points which lie on the
vertical axes at the 0 distance value
correspond, in both figures,
to pairs for which the same language
is compared.
For these points the $D(\alpha, \alpha)
=D_s(\alpha, \alpha)$ are all equal to 0
while the $\Gamma(\alpha, \alpha)$ are positive.
It is easy to see that the self-distances
accounted by the $\Gamma(\alpha, \alpha)$, which compare 
words with different meaning in the same language, 
are, on average, smaller than the $\Gamma(\alpha, \beta)$
which compare words with different meaning in two different languages. 
This fact confirms that the information
carried by $\Gamma(\alpha, \beta)$  is positive.

In other words, closer related languages, not only have
more similar words corresponding to the same meaning,
but the general occurrence and ordering of characters in words 
is more similar.

\begin{figure}
\centering
\includegraphics[height=5cm]{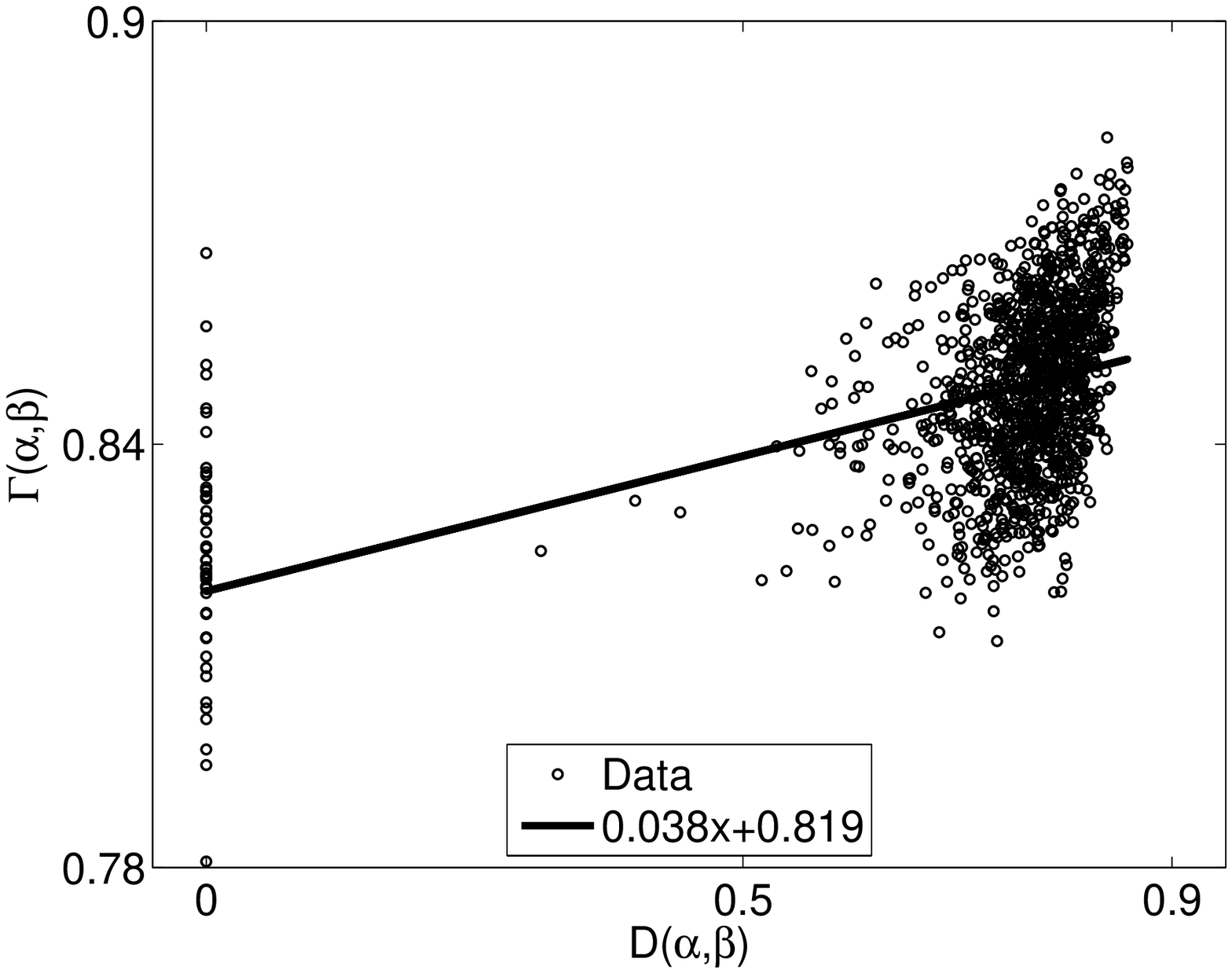}
\includegraphics[height=5cm]{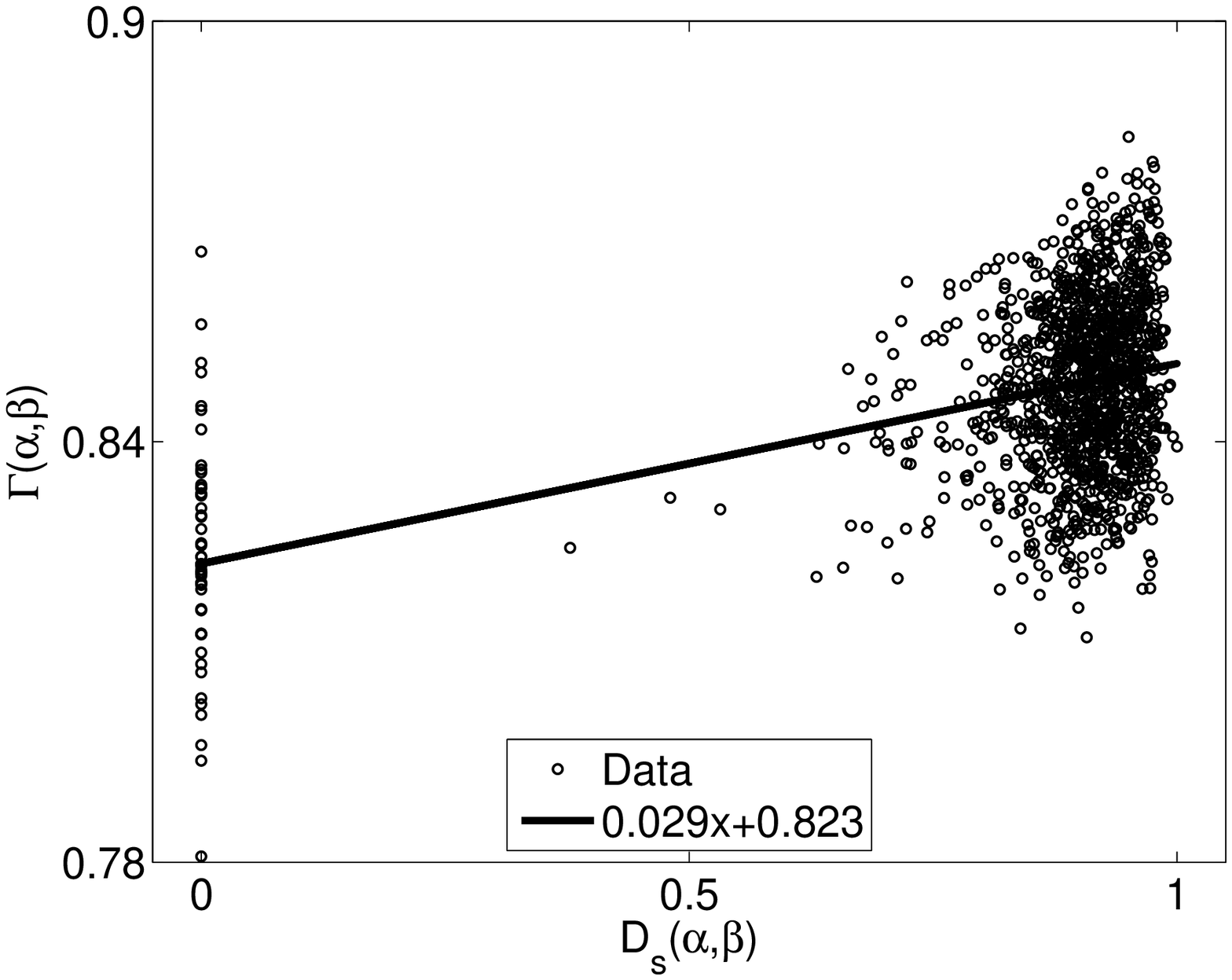}
\caption{Global distance $\Gamma(\alpha,\beta)$ versus 
lexical distance $D(\alpha, \beta)$ (left) and versus 
bi-normalized distance $D_s(\alpha, \beta)$ (right) 
for Austronesian languages. The positive
correlation is evidenced by the best linear fits.
The points which lie on he
vertical axes at the 0 distance value
correspond to pairs for which the same language
is compared.
For these points $D(\alpha, \alpha)=D_s(\alpha, \alpha)$
while $\Gamma(\alpha, \alpha) \neq 0$.}
\label{f2}
\end{figure}

\section{Conclusions}
In this work we have analyzed two different possibilities 
for the definition of automated languages distance. 
More precisely, starting from a Levenshtein distance, 
we have analyzed two possible normalizations. 
The choice between them is only made by using statistical arguments.
 
Our conclusion is that it is 
preferable to use the single normalization 
definition of distance $D(\alpha, \beta)$,
otherwise a part of the information 
about affinities of languages is lost.
In fact, our analysis shows that closer related languages
have smaller global distance.
This means that not only they have more similar words 
for the same meaning,
but the general occurrence and ordering of characters in words 
is more similar.

\section*{Acknowledgments}
We warmly thank S$\o$ren Wichmann for helpful discussion.
We also thank Philippe Blanchard, Luce Prignano and
Dimitri Volchenkov for critical comments on many aspects of the paper.
We are indebted with S.J. Greenhill, R. Blust and R.D.Gray,
for the authorization to use their: 
{\it The Austronesian Basic Vocabulary Database},
http://language.psy.auckland.ac.nz/austronesian
which we consulted in January 2008.

\end{document}